# Training Recurrent Neural Networks as a Constraint Satisfaction Problem

Hamid Khodabandehlou and M. Sami Fadali

*Abstract*—This paper presents a new approach for training artificial neural networks using techniques for solving the constraint satisfaction problem (CSP). The quotient gradient system (QGS) is a trajectory based method for solving the CSP. This study converts the training set of a neural network into a CSP and uses the QGS to find its solutions. The QGS finds the global minimum of the optimization problem by tracking trajectories of a nonlinear dynamical system and does not stop at a local minimum of the optimization problem. Lyapunov theory is used to prove the asymptotic stability of the solutions with and without the presence of measurement errors. Numerical examples illustrate the effectiveness of the proposed methodology and compare it to a genetic algorithm and error backpropagation.

*Keywords* —Neural Networks, Global Optimization, Quotient Gradient System, Modeling, Training

## I. Introduction

After Minsky and Papert showed that two-layer perceptron cannot approximate functions generally [1], it took nearly a decade for researchers to show that multilayer feedforward neural networks are universal approximators [2]. Since then, neural networks have been successfully used in various science and engineering applications [1],[2]. However, training and learning the internal structure of neural networks has remained a challenging problem for researchers.

Training neural networks requires solving a nonlinear non-convex optimization problem and researchers have proposed different approaches to solving it [3]. Classical optimization methods were the first methods used for training neural networks. The most widely used training algorithm is error backpropagation which minimizes an error function using the steepest decent algorithm [4]. Although error backpropagation is easy to implement, it has all the disadvantages of Newton-based optimization algorithms including slow convergence rate and trapping in local minima. Local minima can decrease the generalization ability of the neural network [3],[4].

To cope with these deficiencies, researches proposed other training methods such as supervised learning and global optimization approaches [5],[6],[7]. Supervised learning approaches learn the internal structure of the neural network while learning internal weights of the neural network. Learning the internal structure of the neural network makes these approaches more efficient and less reliant on parameters selected by the user [8],[9].

Researchers proposed different supervised learning methods such as the tiling algorithm, cascade-correlation algorithm, stepnet, and the scaled conjugate algorithm, among others [9]. While in incremental supervised learning approaches network size grows in the training phase which may result in over-fitting, some supervised learning approaches prune the over-fitted network during training [10],[11],[12]. However, few of these methods have been successfully applied to large scale practical problems [9]. This is in contrast to conjugate gradient methods which are attractive for large scale problems due to their fast convergence rate [13]. Quasi-Newton methods are a sophisticated alternative to conjugate gradient methods for supervised learning, although their reliance on exact approximation of the Hessian matrix makes them inefficient in some applications [14].

Global optimization methods are another alternative to cope with deficiencies of Newton-based methods and learn the internal structure of neural networks. Genetic algorithms and simulated annealing have been widely used to train neural networks and optimize network structure [15],[16]. These approaches assume that the quality of the network is related to network topology and parameters. Alopex is another global optimization approach which trains the network using the correlation between changes in weights and changes in the error function. Due to local computations of the Alopex, it is more suitable for parallel computation [17].

Taboo search is another stochastic approach which has been frequently used to train neural networks. It can find the optimal or near optimal solution of the optimization problem [18]. Implementation of taboo search is easier than most global optimization methods and the method is generally applicable to a wide variety of optimization problems [19].

Researchers have used a combination of global optimization methods for training neural networks. GA-SA is a combination of a genetic algorithm and simulated annealing. GA-SA uses a genetic algorithm to make simulated annealing faster to reduce the training time [20],[21],[22]. NOVEL is another hybrid approach which uses a trajectory-based method to find feasible upon work supported by the National Science Foundation under Grant No. IIA-1301726.

H. Khodabandehlou is graduate student at University of Nevada, Reno, Reno, NV, 89557 USA (e-mail: hkhodabandehlou@nevada.unr.edu)

M. Sami Fadali is professor at University of Nevada, Reno, Reno, NV, 89557 USA (e-mail: fadali@unr.edu)

regions of the solution space and then locates local the minima in the feasible regions by local search [23].

Although global optimization methods have been applied for training neural networks, there are other promising global optimization approaches that have not been used for neural network training. Quotient gradient system is a trajectory based method to find feasible solutions of constraint satisfaction problems. QGS searches for the feasible solutions of the CSP along the trajectories of a nonlinear dynamical system [24].

This paper exploits QGS to train artificial neural networks by transforming the training data set into a CSP, then transforms the resulting CSP to an unconstrained minimization problem. After constructing the unconstrained minimization problem, the nonlinear QGS dynamical system is defined. Using the fact that the equilibrium points of the QGS are local minima of the unconstrained minimization problem, a neural network can be trained by integrating QGS over time until it reaches an equilibrium point. The method is easy to implement because constructing the nonlinear dynamical system is similar to deriving the equations of the steepest descent algorithm. The algorithm finds multiple local minima of the optimization by forward and backward integration of the QGS. This provides an easy and straightforward approach to find multiple local minima of the optimization problem. However, like other global optimization methods, finding local minima takes more time than Newton-based methods. Numerical examples show that QGS outperforms error backpropagation and a genetic algorithm and the resulting network has better generalization capability. A preliminary version of the paper which compares the method with error backpropagation was presented in [26].

Solving optimization problems with different initial points is one of the approaches to cope with local minima in Newton-based methods. However, the selected initial points may be in the stability region of the same stable equilibrium point, which makes this approach inefficient. QGS uses backward integration to escape from the stability region of a stable equilibrium point, then enters the stability region of another equilibrium point with forward integration. This allows QGS to explore a bigger region in its search for local minima. The simple implementation, along with the global optimization property of QGS justify its use as a new training method for artificial neural networks.

The remainder of this paper is organized as follows: Section II presents the QGS methodology. Section III describes the structure of the neural network. Application of QGS in training neural network is presented in Section IV. Section V establishes the stability of the proposed method and examines the effect of input errors on its stability. Numerical examples are provided in Section VI and Section VII presents the conclusion.

## II. QUOTIENT GRADIENT SYSTEM

CSP is an active field of research in artificial intelligence and operations research. Lee and Chiang [24], used the trajectories of a nonlinear dynamical system to find the solutions of the CSP. This section reviews their work that forms the basis for our new approach to neural network training which is presented in Section IV.

Consider a system of nonlinear equality and inequality constraints

$$C_I(y) < 0$$
$$C_E(y) = 0, y \in R^{n-l} \qquad (1)$$

To guarantee the existence of the solution of this CSP, $C_I$ and $C_E$ are assumed to be smooth. The CSP can be transformed into the unconstrained optimization problem

$$\min_x f(x) = \frac{1}{2}\|h(x)\|^2, \qquad x = (y, s) \in R^n \qquad (2)$$

$$h(x) = \begin{bmatrix} C_I(y) + \hat{s}^2 \\ C_E(y) \end{bmatrix} \in R^m, \hat{s}^2 = (s_1^2, \ldots, s_l^2)^T \qquad (3)$$

where the slack variable $\hat{s}$ has been introduced to transform the inequality constraints to equality constraints. The global minimum of (2) is the optimal solution of the original CSP. The QGS is a nonlinear dynamical system of equations defined based on the constraint set as

$$\dot{x} = F(x) = -\nabla f(x) := -D_x h(x)^T h(x) \qquad (4)$$

Lee and Chiang showed that stable equilibrium points of the QGS are local minimums of unconstrained minimization problem (2) which are possible feasible solutions of the original CSP. A solution of the QGS starting from initial point $x(0)$ at initial time $t = 0$ is called a trajectory or orbit. An equilibrium manifold is a path connected component of $F^{-1}(0)$. Assuming that $\phi(., x): R \to R^n$ is an orbit of the QGS, an equilibrium manifold $\Sigma$ of the QGS is *stable* if $\forall \epsilon > 0$ there exist $\delta(\epsilon) > 0$ such that

$$x \in B_\delta(\Sigma) \Longrightarrow \phi(t, x) \in B_\epsilon(\Sigma), \forall t \in R \qquad (5)$$

where $B_\delta(\Sigma) = \{x \in R^n: \|x - y\| < \delta, \forall \delta \in R^n \}$. If $\delta$ can be chosen such that

$$x \in B_\delta(\Sigma) \Longrightarrow \lim_{t \to \infty} \phi(t, x) \in B_\epsilon(\Sigma) \qquad (6)$$

the equilibrium manifold is *asymptotically stable*. An equilibrium manifold $\Sigma$ which is not stable is *unstable*. An equilibrium manifold is *pseudo-hyperbolic* if $\forall x \in \Sigma$, the Jacobian of $F(.)$ at $x$ has no eigenvalues with a zero real part on the normal space of $\Sigma$ at $x \in R^n$ and there exist $\epsilon > 0$ such that $\Phi_{-\infty}: B_\epsilon(\Sigma) \to \Sigma$ is locally homeomorphic to projection from $R^n$ to $R^l$ with $l$ the dimension of the equilibrium manifold. The stability region of the stable equilibrium manifold is an open, connected and invariant set and is defined as

$$A(\Sigma_s) = \{x \in R^n : \lim_{t \to \infty} \phi(t, x) \in \Sigma_s\} \qquad (7)$$

The boundary of a stable equilibrium manifold $\Sigma_s$ is the *stability boundary* and is denoted by $\partial A(\Sigma_s)$. QGS is assumed to satisfy the following assumptions.

**Assumptions:** let $\Sigma_s$ be stable equilibrium manifold of QGS
(A1) If an equilibrium manifold $\Sigma$ has nonempty intersection with $\partial A(\Sigma_s)$ then $\Sigma \subset \partial A(\Sigma_s)$
(A2) All the equilibrium manifolds on $\partial A(\Sigma_s)$ are pseudo-hyperbolic and have the same dimension
(A3) The stable and unstable manifolds of equilibrium manifolds on $\partial A(\Sigma_s)$ satisfy the transversality condition.
(A4) The function $H$ satisfies one of the following
  (1) $\|h(x)\|$ is a proper map
  (2) For any $\gamma > 0$ and any closed subset $K$ of

$$\{x \in R^n : \|h(x)\| \leq \gamma, \, Dh(x)^T h(x) \neq 0\},$$
$$\inf\{\|Dh(x)^T h(x)\| : x \in K\} > 0$$

where $Dh(x)$ denotes the gradient of $h(x)$.

The transversality condition of assumption A3 is defined as follows. Let $M_1$ and $M_2$ be manifolds in $R^n$ of codimensions $m_1$ and $m_2$. We say that $M_1$ and $M_2$ intersect transversally if (i) for every $\bar{x} \in M_1 \cap M_2$ there exist an open neighborhood $U_{\bar{x}}$ of $\bar{x}$, and (ii) a system of functions $(h_1, \ldots, h_{m_1})$ for $M_1 \cap U_{\bar{x}}$ and $(\rho_1, \ldots, \rho_{m_2})$ for $M_2 \cap U_{\bar{x}}$ such that the set $\{Dh_i(x), D\rho_j(x), i = 1, \ldots, m_1, j = 1, \ldots, m_2\}$ is linearly independent for all $x \in M_1 \cap M_2 \cap U_{\bar{x}}$ [25]. The following theorem assures the stability of QGS and redefines the stability boundary under assumptions A1-A4.

**Theorem 1** [27]: Let $\sum_s$ be a stable equilibrium manifold of QGS and suppose that assumptions A1-A4 hold. Then we have the following:

1) The QGS is completely stable, i.e., every trajectory of QGS converges to an equilibrium manifold
2) Let $\{\sum_i : i = 1, 2, \ldots\}$ be the set of all equilibrium manifolds on $\partial A(\sum_s)$, then

$$\partial A(\sum_s) = \bigcup_i W^s(\sum_i)$$

where $W^s(\sum)$ is a stable manifold of pseudo-hyperbolic equilibrium manifold and is defined as

$$W^s(\sum) = \{x \in R^n : \lim_{t \to \infty} \phi(t, x) \in \sum\} \quad (8)$$

The next theorem shows that solving the CSP is equivalent to finding stable equilibrium manifolds of the QGS.

**Theorem 2** [24]: Consider the CSP and its associated quotient gradient system. If assumptions (A1-A4) hold, then we have the following

I. Each path component of the solution set of the CSP is a stable equilibrium manifold of the QGS
II. If $\sum$ is a stable equilibrium manifold of the QGS, then $\sum$ consists of non-isolated local minima of the following minimization problem

$$\min_{x \in R^n} V(x) \quad (9)$$

where $V: R^n \to R$ is defined as $V(x) = \frac{1}{2}\|h(x)\|^2$

III. If $n > 2m - 1$ then $\sum$ is a component of the solution set of the CSP if and only if $\sum$ is an $n - m$ dimensional stable equilibrium manifold of the QGS

A stable equilibrium manifold of the QGS may not be in the feasible region of the CSP. In such cases, the QGS must escape from this equilibrium manifold and enter the stability region of another stable equilibrium manifold. If the new equilibrium manifold is not in the feasible region, this process is repeated until the QGS enters the stability region of a feasible equilibrium manifold or until it satisfies a stopping criterion. Once a feasible manifold is reached, QGS is integrated over time until an equilibrium point is reached. To escape from the basin of attraction of a stable equilibrium point, QGS is integrated backward in time until an unstable point is reached. Thus, solving the optimization problem becomes a series of forward and backward integrations of the QGS until the stopping criteria is satisfied.

## III. NEURAL NETWORKS

Function approximation is required in many fields of science and engineering. Neural networks are general function approximators and have been successfully applied to different function approximation applications [2],[3]. Based on the nature of the application, researchers have developed different versions of neural networks such as feedforward networks, recurrent neural networks, liquid state networks and wavelet networks among the others [28].

In this study, we consider a three-layer fully recurrent neural network with smooth activation functions. Fig. 1 illustrates the internal structure of the neural network. The network has input $u(k) = [u_1(k) \ldots u_n(k)]^T$, internal state $z(k) = [z_1(k) \ldots z_m(k)]^T$ and output $\hat{y}(k) = [\hat{y}_1(k) \ldots \hat{y}_t(k)]^T$. The input-output equation of the network is described as

$$z(k) = \psi(Wu(k) + Sz(k-1))$$
$$\hat{y}(k) = V z(k) \quad (10)$$

$W$, $S$ and $V$ are network weights matrices whose size is dependent on the number of network inputs, outputs and hidden layer nodes. For a network with $n$ inputs, $t$ outputs and $m$ hidden layer nodes, $W \in R^{m \times n}$, $S \in R^{m \times m}$ and $V \in R^{t \times m}$. The cost function for training neural network is the traditional sum of squared errors (SSE)

$$SSE = \sum_{k=1}^{N} e(k)^T e(k)$$
$$= \sum_{k=1}^{N} (\hat{y}(k) - y(k))^T (\hat{y}(k) - y(k)) \quad (11)$$

where $\hat{y}(k)$ is the network output, $y(k)$ is the measured output, and $N$ is the total number of training samples.

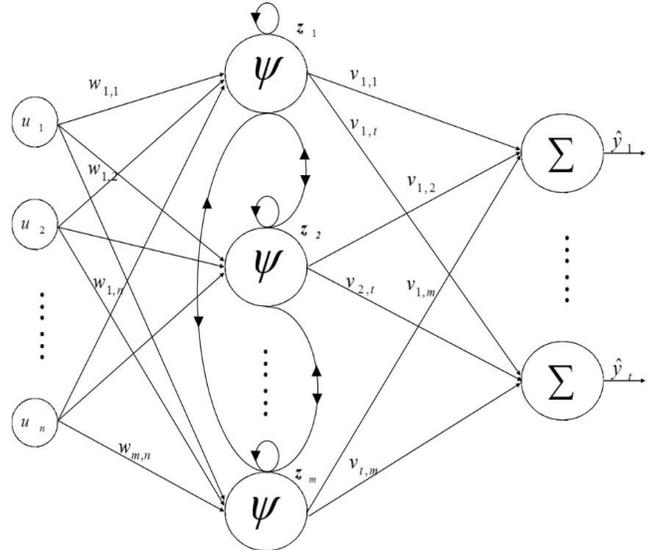

Fig. 1. Artificial neural network structure

## IV. APPLYING QGS TO NEURAL NETWORK TRAINING

Solving the CSP is equivalent to an unconstrained minimization problem (2). QGS is a trajectory-based method to find the local minima of (2) which are the possible feasible solutions of the CSP. To train neural networks using QGS, we consider the training set as equality constraints of the CSP and then transform the CSP into unconstrained minimization problem as (2) and then we use the second part of Li and Chiang's work which is equilibrium points of QGS are local minimums of unconstrained minimization problem. If $N$ measurements are available, the CSP can be written as

$$\begin{aligned} \boldsymbol{h}(\boldsymbol{x}) &= [h_i(\boldsymbol{x})], i = 1,2,\dots,N \\ h_i(\boldsymbol{x}) &= V\boldsymbol{\psi}\big(W\boldsymbol{u}(i) + S\boldsymbol{z}(i-1)\big) - y(i) \end{aligned} \quad (12)$$

The network state vector $\boldsymbol{x}$ includes all the network parameters, i.e., all entries of $V, W$ and $S$. More specifically, if we partition $V, W$ and $S$ as

$$V = \begin{bmatrix} \boldsymbol{v}_1^{\mathrm{T}} \\ \vdots \\ \boldsymbol{v}_m^{\mathrm{T}} \end{bmatrix}_{t \times m} W = \begin{bmatrix} \boldsymbol{w}_1^{\mathrm{T}} \\ \vdots \\ \boldsymbol{w}_m^{\mathrm{T}} \end{bmatrix}_{m \times n} S = \begin{bmatrix} \boldsymbol{s}_1^{\mathrm{T}} \\ \vdots \\ \boldsymbol{s}_m^{\mathrm{T}} \end{bmatrix}_{m \times m} \quad (13)$$

then $\boldsymbol{x}$ is defined as

$$\begin{aligned} \boldsymbol{x} &= [x_i]_{n_p \times 1} = [\boldsymbol{v}_1, \dots, \boldsymbol{v}_m, \boldsymbol{w}_1, \dots, \boldsymbol{w}_m, \boldsymbol{s}_1, \dots, \boldsymbol{s}_m]^{\mathrm{T}} \\ n_p &= m^2 + m \times (n + t) \end{aligned} \quad (14)$$

Since the training set does not contain any inequality constraints, slack variables are not needed. Using the training set, the QGS for training the neural network can be defined as

$$\dot{\boldsymbol{x}} = -\boldsymbol{f}(\boldsymbol{x}) = -D_x \boldsymbol{h}(\boldsymbol{x})^{\mathrm{T}} \boldsymbol{h}(\boldsymbol{x}) \quad (15)$$

where

$$D_x \boldsymbol{h}(\boldsymbol{x}) = \begin{bmatrix} \dfrac{\partial h_1(\boldsymbol{x})}{\partial \boldsymbol{x}} \\ \vdots \\ \dfrac{\partial h_N(\boldsymbol{x})}{\partial \boldsymbol{x}} \end{bmatrix}_{N \times n_p} \quad (16)$$

To train neural network using QGS, we use the fact that the equilibrium points of QGS are local minima of the unconstrained minimization problem. Therefore the algorithm needs to find an equilibrium point of QGS and then escape from that equilibrium point and move toward another equilibrium point of QGS. The first step is to integrate the QGS from a starting point, which need not be feasible, to find an equilibrium point. Next, we escape from the stability region of the stable equilibrium point to an unstable point with backward integration of QGS in time. The eigenvalues of the Jacobian matrix can be used as a measure of stability and instability. The algorithm continues until it cannot find any new equilibrium point or until it satisfies the stopping criterion.

Because neural network training has equilibrium points which can be considered as zero-dimensional equilibrium manifolds, assumptions A1, A2 and A3 hold. When the activation function of the neural network is a one-one invertible function, $\|\boldsymbol{h}(\boldsymbol{x})\|$ is a proper map. Assumption 4 also holds because the QGS is asymptotically stable and $\|\boldsymbol{h}(\boldsymbol{x})\|$ is proper.

## V. STABILITY

Any training algorithm must be stable, even in presence of measurement error and uncertainties. We use Lyapunov stability theory to prove the asymptotic stability of equilibrium points and their asymptotic stability in the presence of measurement errors.

**Theorem 3:** The equilibrium points of the quotient gradient system are asymptotically stable

*Proof:* Consider the Lyapunov function

$$V(\boldsymbol{x}) = \boldsymbol{h}^{\mathrm{T}}(\boldsymbol{x})\boldsymbol{h}(\boldsymbol{x}) \quad (17)$$

$V(\boldsymbol{x})$ is a locally positive definite function of the state that is equal to zero at global optima of the optimization problem. Thus, $V(\boldsymbol{x})$ is a locally positive definite function in the vicinity of each equilibrium point. The derivative of the Lyapunov function along the system trajectories is

$$\dot{V} = \left(\frac{\partial V}{\partial \boldsymbol{x}}\right)^{\mathrm{T}} \dot{\boldsymbol{x}} = -\boldsymbol{h}^{\mathrm{T}} D\boldsymbol{h} D\boldsymbol{h}^{\mathrm{T}} \boldsymbol{h} = -\|D\boldsymbol{h}^{\mathrm{T}}\boldsymbol{h}\|^2 \quad (18)$$

The derivative of the Lyapunov function is negative definite in the vicinity of each equilibrium point of the QGS, i.e. in the vicinity of each local minimum of the optimization problem. The Jacobian $D\boldsymbol{h}$ is positive definite in the vicinity of the equilibrium points because they are minima of the cost function. Therefore, all the equilibrium points of the QGS are locally asymptotically stable. ●

Under certain conditions, the equilibrium points are exponentially stable as shown in the next theorem.

**Theorem 4:** The equilibrium points of the QGS are exponentially stable.

*Proof:* Consider the Lyapunov function of (17). When there is no repeated measurement, $D\boldsymbol{h}$ is full rank and therefore $D\boldsymbol{h} D\boldsymbol{h}^{\mathrm{T}}$ is a positive definite matrix. Assume that $\sigma_{\min}$ is the smallest singular value of the positive definite matrix $D\boldsymbol{h} D\boldsymbol{h}^{\mathrm{T}}$. The derivative of the Lyapunov function can be written as

$$\begin{aligned} \dot{V} &= \left(\frac{\partial V}{\partial \boldsymbol{x}}\right)^{\mathrm{T}} \dot{\boldsymbol{x}} = -\boldsymbol{h}^{\mathrm{T}}(\boldsymbol{x}) D\boldsymbol{h} D\boldsymbol{h}^{\mathrm{T}} \boldsymbol{h}(\boldsymbol{x}) \\ &\leq -\sigma_{\min} \boldsymbol{h}^{\mathrm{T}}(\boldsymbol{x}) \boldsymbol{h}(\boldsymbol{x}) \\ &= -\sigma_{\min} \|\boldsymbol{h}(\boldsymbol{x})\|^2 \end{aligned} \quad (19)$$

where $\sigma_{\min}$ is the smallest eigenvalue of Therefore $\dot{V} \leq -\sigma_{\min} V$ and the equilibrium points of the QGS are exponentially stable. With the bounded input and output assumption, (29) yields that $\|D\boldsymbol{h}\|_2$ is bounded. Therefore the spectral radius and consequently smallest singular value of $D\boldsymbol{h}$ are finite.

Measurement errors and noise can make the measurements inaccurate and destabilize the system. Fortunately, QGS can tolerate relatively large measurement error. In neural networks, measurement errors lead to errors in neural network inputs. Consider the QGS as a function of $\boldsymbol{x}$ and $\boldsymbol{u}$, i.e, $\dot{\boldsymbol{x}} = -\boldsymbol{f}(\boldsymbol{x}, \boldsymbol{u})$ and let the measurement errors change $\boldsymbol{u}$ to $\boldsymbol{u} + \Delta \boldsymbol{u}$. Assuming that $\Delta \boldsymbol{u}$ is small

$$\begin{aligned} \dot{\boldsymbol{x}} &= -\boldsymbol{f}(\boldsymbol{x}, \boldsymbol{u} + \Delta \boldsymbol{u}) \\ &= -\boldsymbol{f}(\boldsymbol{x}, \boldsymbol{u}) - \frac{\partial \boldsymbol{f}(\boldsymbol{x}, \boldsymbol{u})}{\partial \boldsymbol{u}} \Delta \boldsymbol{u} + HOT \end{aligned} \quad (20)$$

where *HOT* denotes higher order terms. For sufficiently small $\Delta u$, we can neglect higher order terms and write

$$\dot{x} = -f(x, u + \Delta u) \cong -f(x, u) - \frac{\partial f(x, u)}{\partial u} \Delta u \qquad (21)$$
$$= -f(x, u) + g(x, u, \Delta u)$$

Assuming that the activation functions of the neural network are continuously differentiable, $g$ will be continuously differentiable and hence $g$ is Lipschitz for all $t > 0$ and $\in T \subset R^n$, with $T$ the domain that contains the equilibrium point. Assume that the perturbation term satisfies the linear growth bound

$$\|g(x, u, \Delta u)\| \leq \gamma \|x\|, \forall t \geq 0, \forall x \in T \qquad (22)$$

To find a bound on the perturbation that guarantees stability, we need the following property of matrix norms

**Fact:** For every $A: C^n \to C^m$

$$\|A\|_2 \leq \|A\|_F \leq \sqrt{r} \|A\|_2 \qquad (23)$$

where $\|A\|_F$ is the Frobenius norm of $A$ and $r$ is its rank.

**Theorem 5:** Assume that the input and output of the network are bounded, $\|y\| \leq K_y$ and $\|u\| \leq K_u$, and the corresponding neural network has $n$ inputs, $m$ hidden layer nodes and we have $N$ measurements. The equilibrium of the perturbed QGS is asymptotically stable if

$$\gamma < N\sqrt{Nm} \left[ \left( \sqrt{m} + \sqrt{K_y(\sqrt{n}K_u + m)} \right)^2 \right] \qquad (24)$$

**Proof:** Consider the Lyapunov function $V(x) = h^T(x)h(x)$. The derivative of $V(x)$ including the perturbation is

$$\dot{V} = (-h^T Dh + g^T)Dh^T h \qquad (25)$$
$$= -\|Dh^T h\|^2 + g^T Dh^T h$$

For a negative definite $\dot{V}$, we need

$$g^T Dh^T h < \|Dh^T h\|^2 \qquad (26)$$

This condition is satisfied if

$$\|g\| < \|Dh^T h\| \leq \|h\| \times \|Dh\| \qquad (27)$$

Using the nonlinearity of (12) with a bounded output, $h$ satisfies

$$\|h\| \leq N(m\|x\| + K_y) \qquad (28)$$

The Jacobian of the hyperbolic function gives

$$\|Dh\|_F \leq \sqrt{Nm}(1 + \sqrt{n}\|u\|\|x\| + m\|x\|) \qquad (29)$$

Using proposition 1 gives the 2-norm bound

$$\|Dh\|_2 \leq \sqrt{Nm}(1 + \sqrt{n}\|u\|\|x\| + m\|x\|) \qquad (30)$$

By combining (29), (27), (26) and (22)

$$\gamma\|x\| < N\sqrt{Nm}(1 + \sqrt{n}\|u\|\|x\| + m\|x\|)(m\|x\| + K_y) \qquad (31)$$

Using the input bound $\|u\| \leq K_u$ gives the condition for negative definite $\dot{V}$

$$\gamma < N\sqrt{Nm} \left[ \left( \sqrt{m} + \sqrt{K_y(K_u\sqrt{n} + m)} \right)^2 \right] \qquad (32)$$

∎

## VI. SIMULATION RESULTS

To illustrate the effectiveness of QGS for training neural networks, we use a QGS trained network for nonlinear system identification and compare the results with a genetic algorithm we use the results of [26]. The genetic algorithm optimization uses the MATLAB optimization toolbox with a population size of 10000, Roulette selection, adaptive feasible mutation, scattered crossover, and top fitness scaling to get the best results.

### A. Example 1: Nonlinear system

Our first benchmark system is the second order nonlinear system chosen from [31]. The input-output equation of the system is described as

$$y(k+1) = \frac{y(k)y(k-1)(y(k) + 0.25)}{1 + y(k)^2 + y(k-1)^{\wedge}2} + q(k) \qquad (33)$$

$q(k)$ is the system input and $y(k)$ is the system output. $q(k)$ is zero-mean normally distributed with standard deviation $\sigma = 0.5$. The input to the neural network is $u(k) = [q, q(k-1), y(k), \ldots, y(k-1)]^T$ and $y(k+1)$ is the target output for training. All the initial network parameter values are zero-mean normally distributed with standard deviation $\sigma = 0.5$. The optimal number of hidden layer nodes is found to be $m = 8$ and the total number of training sets is $N = 200$. The activation function of the neural network is the tangent hyperbolic function

$$\psi(x) = \tanh(x) = \frac{e^x - e^{-x}}{e^x + e^{-x}} \qquad (34)$$

After initializing with random initial values, QGS finds 47 local minima of the optimization problem. The local minimum with the best generalization capability is the global minimum or close to the optimal solution of the optimization problem.

Table I summarizes the mean squared error (MSE) for test data for QGS network, genetic algorithm network, and error backpropagation network. The MSE of the QGS network is less than the MSE for the genetic algorithm network and both networks outperform the backpropagation trained network. Other simulation results that are not included here for brevity, including generalization errors, demonstrate that back propagation gives much worse results than the two other networks. Hence, we do not include back propagation in the remainder of this example.

Table I. Mean squared error

| Training method | QGS | GA | EBP |
|---|---|---|---|
| MSE | 0.00797 | 0.0082 | 0.0187 |

Fig. 2 shows the outputs of the system, the QGS trained network, and the genetic algorithm trained network and Fig. 3

shows the same outputs for test data. While the training results is the same for both networks, Fig. 3 shows that QGS trained network has better generalization performance on random input as test data and has smaller generalization error.

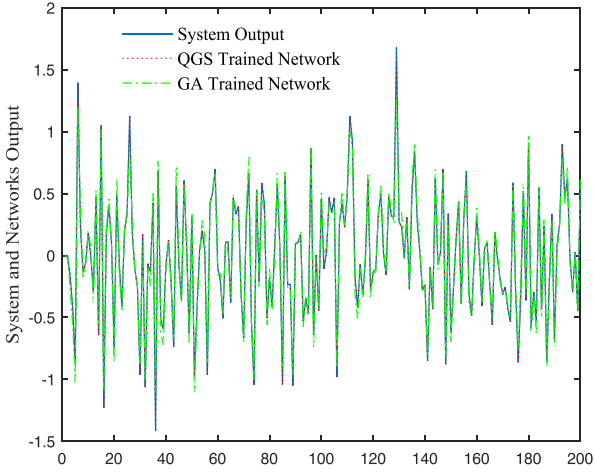

Fig. 2. Outputs of the system and the trained neural networks for the training set

Fig. 4 shows the generalization error for the QGS trained network and genetic algorithm trained network. While both networks have similar performance with the training data as input, Fig. 4 illustrates that the QGS trained network has better generalization capability in terms of maximum generalization error percentage and mean squared error for test data. The average absolute generalization error of QGS trained network is 1.05% while average absolute generalization error of genetic algorithm trained network is 1.38%.

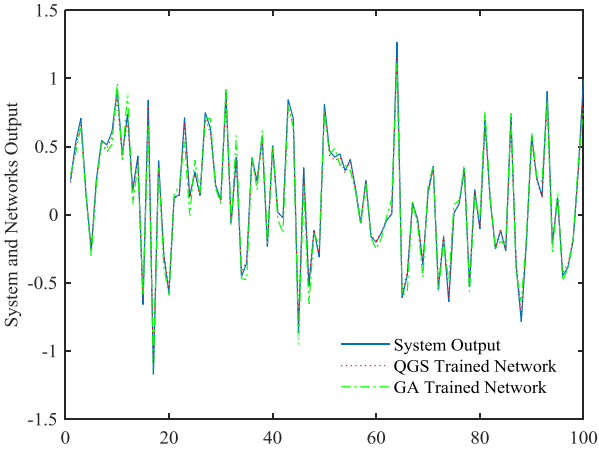

Fig. 3. Outputs of the system and the trained neural networks for test data

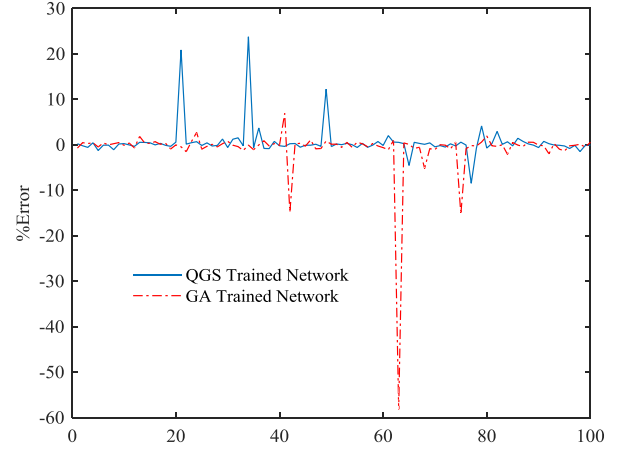

Fig. 4. Generalization Error

*B. Example 2: NARMA System*

Our first benchmark system is a tenth order nonlinear autoregressive moving average (NARMA) process [29]. The input-output equation of the system is described as

$$y(k+1) = 0.3y(k) + .05y(k)\sum_{i=1}^{9}y(k-i) \quad (35)$$
$$+1.5 \times q(k-9)q(k) + 0.1$$

$q(k)$ is the system input and $y(k)$ is the system output. $q(k)$ is zero-mean normally distributed with standard deviation $\sigma = 0.5$. The input to the system is $\boldsymbol{u}(k) = [q(k), \ldots, q(k-9), y(k), \ldots, y(k-4)]^T$ and $y(k+1)$ is the target output for training. All the initial network parameter values are zero-mean normally distributed with standard deviation $\sigma = 0.5$. The optimal number of hidden layer nodes is found to be $m = 6$ and the total number of training sets is $N = 100$. After initializing with random initial values, QGS finds 36 local minima for the optimization problem. The local minimum with the best generalization capability is the global minimum or close to optimal solution of the optimization problem.

Table II. summarizes the MSE for test data for the QGS trained network, the genetic algorithm trained network and the error backpropagation trained network. The MSE of QGS is smaller than the MSE for the genetic algorithm trained network. Both networks outperform the error backpropagation trained network. As in Example 1, other simulation results are much worse for back propagation than for the other two networks and we omit back propagation results for the remainder of this example.

Table II. Mean squared error

| Training method | QGS | GA | EBP |
|---|---|---|---|
| MSE | 0.0026 | 0.0038 | 0.0087 |

Fig. 5 shows the outputs of the system, the QGS trained network, and the genetic algorithm trained network. Fig. 6 shows the same outputs for test data. Fig. 5 shows that QGS trained network has better performance on train data and Fig. 6 shows that QGS trained network outperforms genetic algorithm trained network on random input as a test data.

Fig. 7 shows the generalization error for the QGS trained network and for the genetic algorithm trained network. While both networks have similar performance with the training data as input, Fig. 6 and Fig. 7 show that the QGS trained network has better generalization capability. The average absolute generalization error of QGS trained network is 1.45% while average absolute generalization error of genetic algorithm trained network is 2.86%.

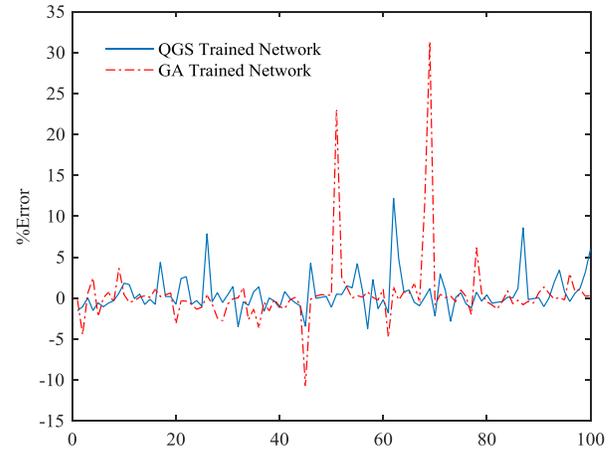

Fig. 7. Generalization Error

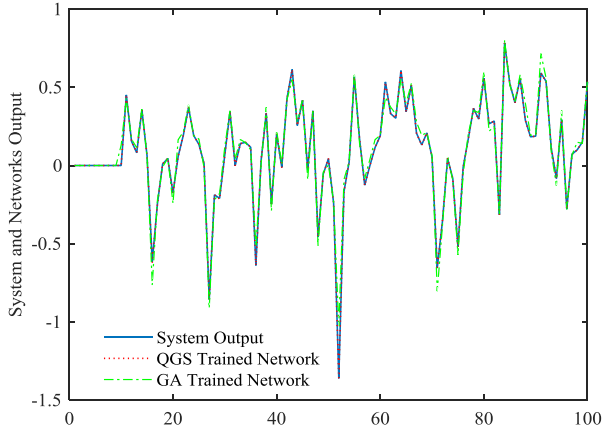

Fig. 5. Outputs of the system and the trained neural networks for the training set

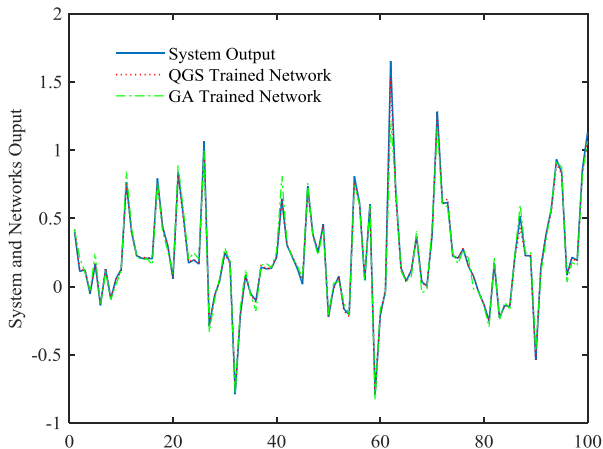

Fig.6. Outputs of the system and the trained neural network with the test data as input.

### VII. CONCLUSION

In this study, we introduce a new training algorithm for neural network using the QGS. QGS uses trajectories of a nonlinear dynamical system to find a local minima of the optimization problem. The local minimum with the best generalization capability is the global minimum of the optimization problem. Simulation results shows that QGS trained network performs better than networks trained using genetic algorithm and error backpropagation. In particular, QGS networks have better generalization properties, faster training time in comparison to genetic algorithm and are more robust to errors in the inputs.

In contrast to Newton based methods, QGS does not need multiple initial values to find multiple local minima and does not need a huge number of measurements for training. Therefore, QGS is particularly suited to applications with a limited number of available input-output measurements. Future work will exploit the projected gradient system (PGS) [30], together with the QGS, to develop a training algorithm for neural networks by searching for local minima of an optimization problem.